\documentclass[]{ceurart}

\begin{document}

\copyrightyear{2025}
\copyrightclause{Copyright for this paper by its authors.
  Use permitted under Creative Commons License Attribution 4.0
  International (CC BY 4.0).}

\conference{CLEF 2025 Working Notes, 9 -- 12 September 2025, Madrid, Spain}

\title{Unibuc - NLP at "Voight-Kampff" Generative AI Detection PAN 2025}
\title[mode=sub]{Notebook for PAN at CLEF 2025}

\author[1, 3]{Teodor Marchitan}[%
email=teodor.marchitan@s.unibuc.ro,
]
\address[1]{Faculty of Mathematics and Computer Science, University of Bucharest, Romania}

\author[2, 3]{Claudiu Creanga}[%
email=claudiu.creanga@s.unibuc.ro,
]
\address[2]{Interdisciplinary School of Doctoral Studies, University of Bucharest, Romania}

\author[1, 3]{Liviu P. Dinu}[%
email=ldinu@fmi.unibuc.ro,
]
\address[3]{HLT Research Center, Bucharest, Romania}

\begin{abstract}
  This paper describes the approach of the Unibuc - NLP team in tackling the "Voight-Kampff" Generative AI Detection task from PAN at CLEF 2025 \cite{bevendorff:2025a, bevendorff:2025b}. In the Subtask 1: Voight-Kampff AI Detection Sensitivity we have explored different approaches of leveraging embeddings extracted from LLMs, using either classical ML models or further fine-tuning LLMs with a classification head for the downstream binary classification task. Our experiments showed that using a majority vote ensemble of classical ML models (Light GBM, XGB, Logistic Regression and SVM) is comparable with fine-tuning for downstream task achieving around $99.4\%$ F1 score. For the Subtask 2: Human-AI Collaborative Text Classification we explored a way of combining features at different layers extracted using transformer backbone with layer-wise projection and attentive pooling which proved modest results.
\end{abstract}


\maketitle

\section{Introduction}

As artificial intelligence continues to advance and generative AI technologies are more and more popular, distinguishing between machine-generated and human-authored texts has become very difficult. The "Voight-Kampff" Generative AI Detection task from PAN at CLEF 2025 \cite{bevendorff:2025a, bevendorff:2025b} aims to advance the research in the field and find ways to develop real systems capable of detecting not only AI generated texts but also machine-altered ones. The need for such systems is becoming a necessity day by day in order to maintain the intergrity and authenticity of the content from the internet. Such tools can help mitigate the risks associated with social manipulation, deepfakes, fake news, propaganda, misinformation and so on.

For the Subtask 1: Voight-Kampff AI Detection Sensitivity we experimented various methods of utilizing the embeddings from large language models (LLMs), including classical machine learning approaches and fine-tuning classification head for the binary classification task. The results indicate that a majority vote ensemble of classical models - specifically LightGBM, XGBoost, Logistic Regression and SVM - offers a slightly better performance and faster training times and resources compared to the fine-runing approach.

For the Subtask 2: Human-AI Collaborative Text Classification, we have tried a different approach by fusing together layers of a transformer backbone model with layer-wise projection and attentive pooling strategies.

\section{Related work}

The detection of AI-generated text has become a rapidly evolving research area, particularly with the advent of large language models. Recent shared tasks such as the PAN 2024 "Voight-Kampff" Generative AI Authorship Verification challenge have provided valuable benchmarks and insights into the capabilities and limitations of current detection systems \cite{froebe:2023b, bevendorff:2024}. Early approaches to authorship verification, such as one-class classification \cite{10.1145/1015330.1015448} and compression-based models \cite{10.1145/3098954.3104050} have been extended to address the challenges posed by neural text generation \cite{uchendu-etal-2020-authorship, uchendu-etal-2021-turingbench-benchmark}. More recent work has focused on zero-shot detection methods, leveraging statistical properties of model outputs \cite{hans2024spottingllmsbinocularszeroshot}. Despite these advances, studies have shown that both human and automated heuristics can be unreliable, especially as generative models improve in mimicking human style \cite{Jakesch_2023, sadasivan2025aigeneratedtextreliablydetected}. Our work builds on these foundations, evaluating both classical and transformer-based approaches for robust detection of machine-generated and collaboratively authored texts.

\section{Dataset}

\subsection{Subtask 1: Voight-Kampff AI Detection Sensitivity}

The first subtask is formulated as a binary classification problem and the dataset contains either human-written texts (label $0$) or ai-written texts (label $1$). The genre of the text is also available and it can be one of the following: essays, news or fiction. Some of the news texts were sampled from the last year task dataset \cite{bevendorff2024voight}.

\begin{figure}[htbp]
    \centering
    \includegraphics[width=1\linewidth]{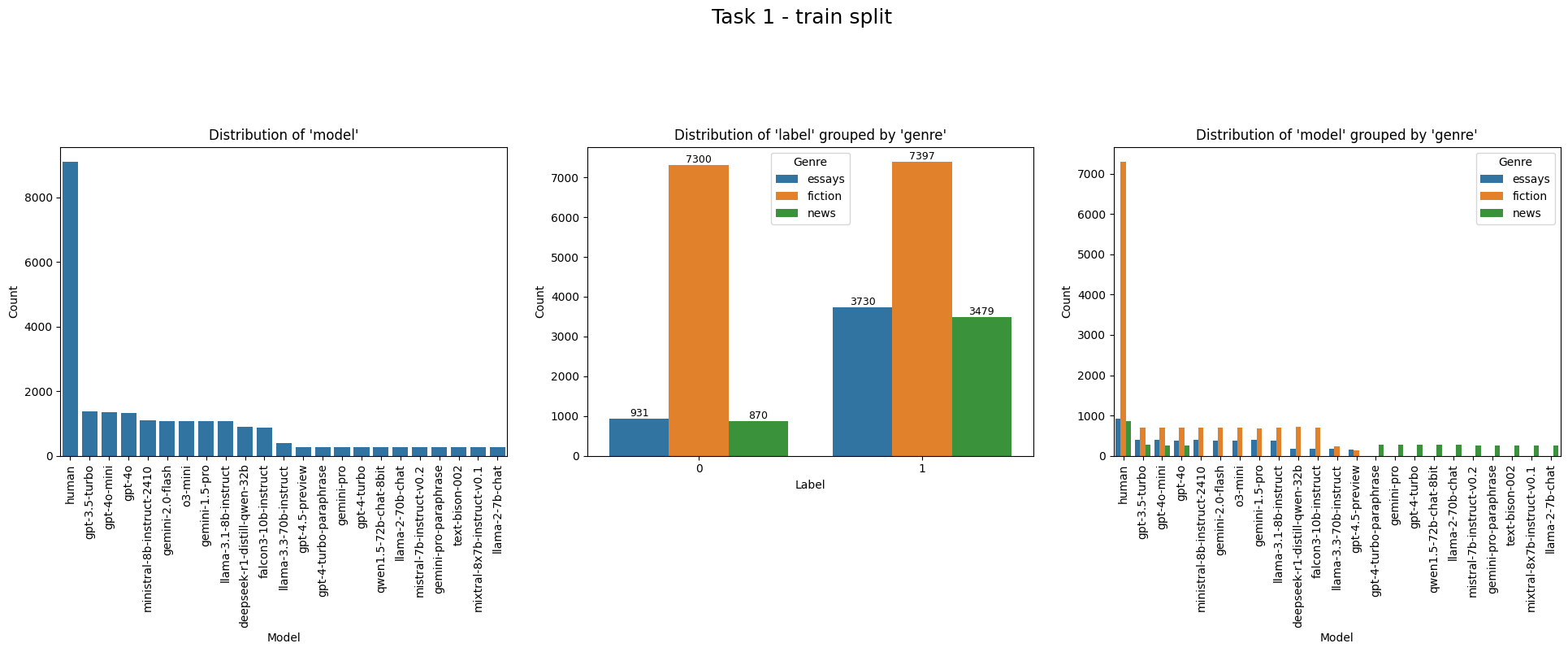}
    \caption{Task 1 dataset overview - train split}
    \label{fig:task1-train-overview}
\end{figure}

\begin{figure}[htbp]
    \centering
    \includegraphics[width=1\linewidth]{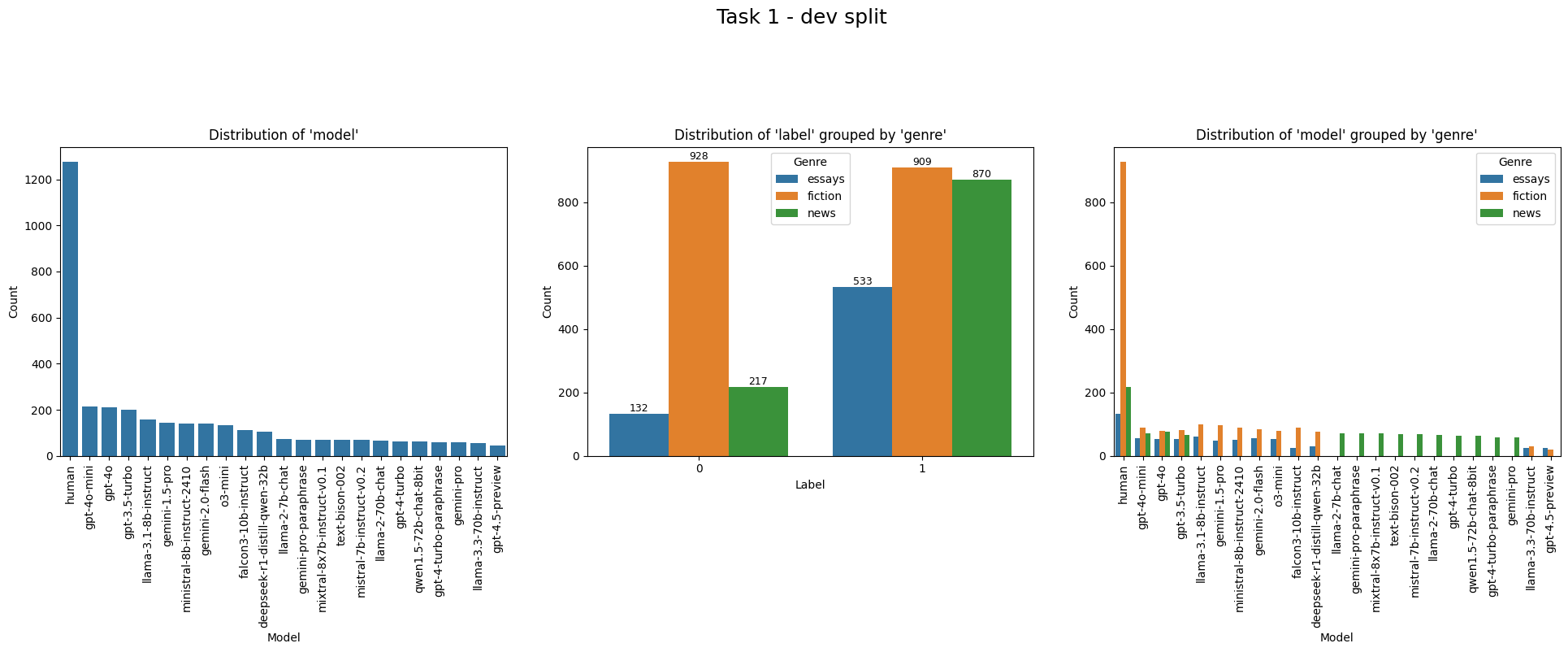}
    \caption{Task 1 dataset overview - dev split}
    \label{fig:task1-dev-overview}
\end{figure}

Overall the train split has $23,707$ samples and dev split has $3,589$. The distribution of the labels in each split is shown in \autoref{table:task1-label-distribution}. The predominant genre for both labels is fiction, followed by essays and news. As for the models used to generate text, gpt \cite{openai2024gpt4technicalreport} comes first followed by mistral \cite{jiang2023mistral7b}, gemini \cite{geminiteam2025geminifamilyhighlycapable}, llama \cite{touvron2023llamaopenefficientfoundation} and deepseek \cite{deepseekai2025deepseekr1incentivizingreasoningcapability} (more info in \autoref{fig:task1-train-overview} and \autoref{fig:task1-dev-overview}).

\begin{table}[h]
    \caption{Label distribution in train and dev splits for task 1 dataset.}
    \label{table:task1-label-distribution}
    
  \begin{tabular}{c |cc | c}
    \toprule
      & \texttt{human-written (0)} & \texttt{ai-written (1)} & \texttt{Total} \\
    \midrule
    \texttt{Train} & $9,101$ & $14,606$ & $23,707$ \\
    \texttt{Dev} & $1,277$ & $2,312$ & $3,589$ \\
    \bottomrule
  \end{tabular}
\end{table}

\subsection{Subtask 2: Human-AI Collaborative Text Classification}

The second subtask is a multi-class classification problem where the goal is to classify texts into six distinct categories based on the nature of human and machine contributions:

\begin{itemize}
    \item \textbf{Fully human-written:} The document is entirely authored by a human without any AI assistance.
    \item \textbf{Human-initiated, then machine-continued:} A human starts writing, and an AI model completes the text.
    \item \textbf{Human-written, then machine-polished:} The text is initially written by a human but later refined or edited by an AI model.
    \item \textbf{Machine-written, then machine-humanized (obfuscated):} An AI generates the text, which is later modified to obscure its machine origin.
    \item \textbf{Machine-written, then human-edited:} The content is generated by an AI but subsequently edited or refined by a human.
    \item \textbf{Deeply-mixed text:} The document contains interwoven sections written by both humans and AI, without a clear separation.
\end{itemize}

Overall the train split has $288,919$ samples and dev split has $72,661$. The distribution of the labels in each split is shown in \autoref{table:task2-label-distribution}.

\begin{figure}[htbp]
    \centering
    \includegraphics[width=1\linewidth]{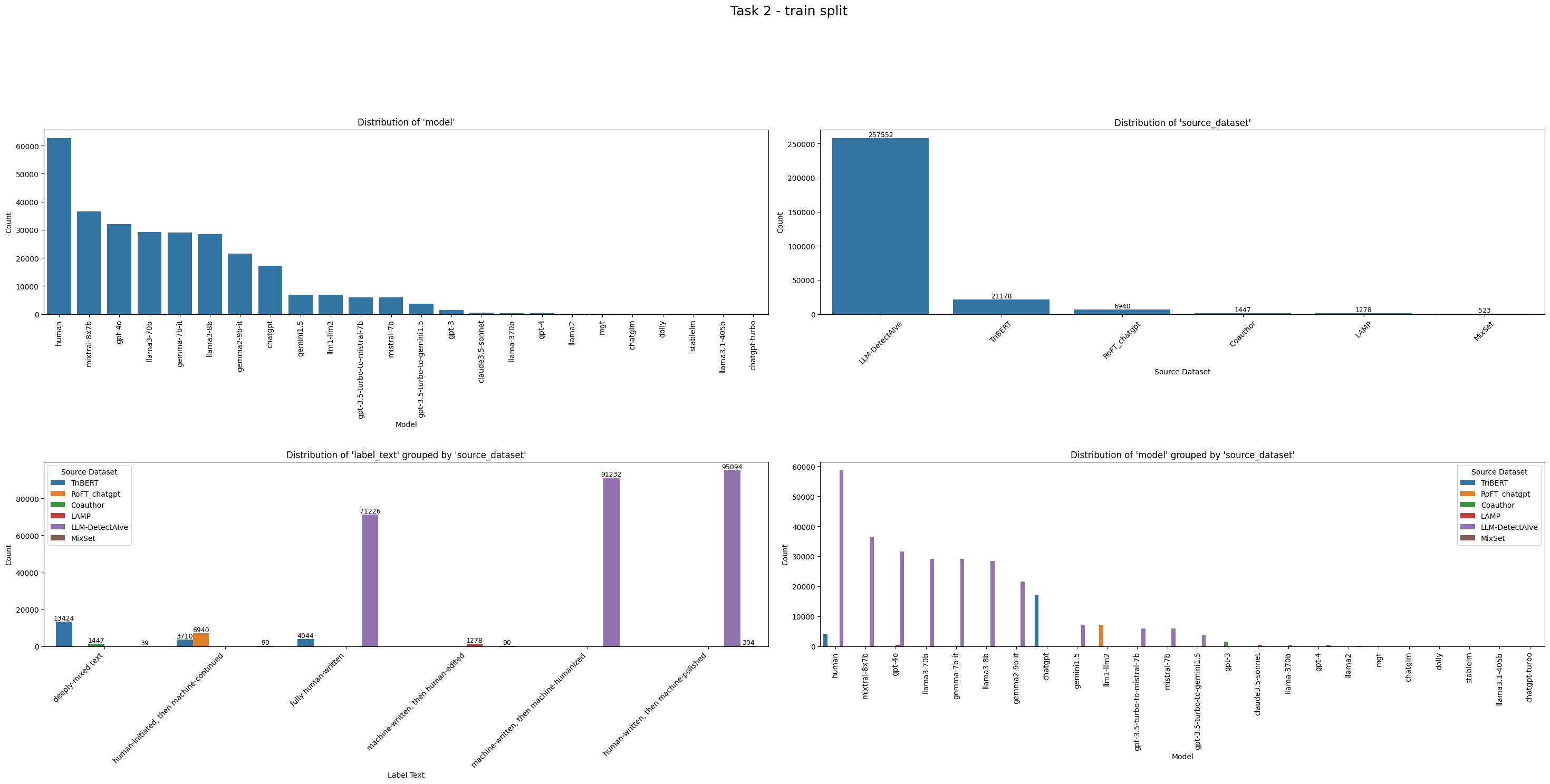}
    \caption{Task 2 dataset overview - train split}
    \label{fig:task2-train-overview}
\end{figure}

In the train split, models used for text rewriting and generation are mostly from mistral \cite{jiang2023mistral7b}, gpt \cite{openai2024gpt4technicalreport}, gemini \cite{geminiteam2025geminifamilyhighlycapable} and llama \cite{touvron2023llamaopenefficientfoundation} (more info in \autoref{fig:task2-train-overview}). Most of the texts that are fully human-written, machine-written then machine-humanized or human-written then machine-polished come from the LLM-DetectAlve dataset \cite{abassy-etal-2024-llm}. The next big source of the texts is TriBERT \cite{zeng2023automaticboundarydetectionhumanai} where ChatGPT is mainly used to produce deeply-mixed texts and human-initiated then machine-continued.

\begin{figure}[htbp]
    \centering
    \includegraphics[width=1\linewidth]{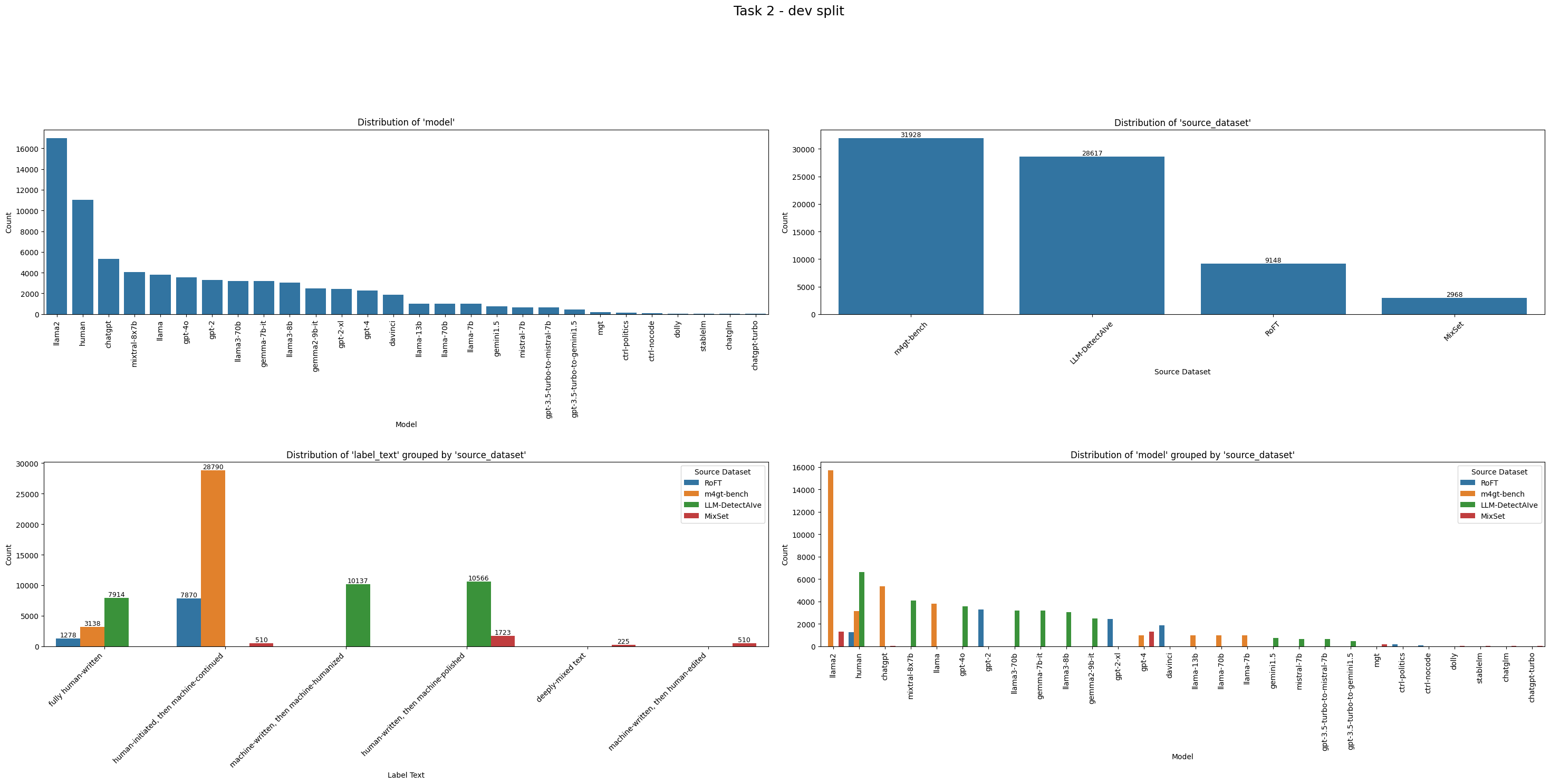}
    \caption{Task 2 dataset overview - dev split}
    \label{fig:task2-dev-overview}
\end{figure}

On the other hand, in the dev split llama \cite{touvron2023llamaopenefficientfoundation} is mainly used for text generation followed by gpt \cite{openai2024gpt4technicalreport} and mistral \cite{jiang2023mistral7b}, gemini \cite{geminiteam2025geminifamilyhighlycapable} being less used in this split. While in this split a good amount of fully human-written, machine-written then machine-humanized or human-written then machine-polished texts come from the LLM-DetectAlve dataset \cite{abassy-etal-2024-llm}, there is a huge amount of human-written and human-initiated then machine-continued texts that are sampled from m4gt-bench \cite{wang2024m4gtbenchevaluationbenchmarkblackbox}. Also in dev set there are more samples from RoFT \cite{kushnareva2024aigeneratedtextboundarydetection} than in the train set (more information in \autoref{fig:task2-dev-overview}).

\begin{table}[h]
    \caption{Label distribution in train and dev splits for task 2 dataset.}
    \label{table:task2-label-distribution}
    
  \begin{tabular}{l |cc}
    \texttt{Label category} & \texttt{Train} & \texttt{Dev} \\
    \midrule
    \texttt{Fully human-written (0)} & $14,910$ & $225$ \\
    \texttt{Human-initiated, then machine-continued (3)} & $1,368$ & $510$ \\
    \texttt{Human-written, then machine-polished (1)} & $10,740$ & $37,170$ \\
    \texttt{Machine-written, then machine-humanized (obfuscated) (2)} & $75,270$ & $12,330$ \\
    \texttt{Machine-written, then human-edited (5)} & $95,398$ & $12,289$ \\
    \texttt{Deeply-mixed text (4)} & $91,232$ & $10,137$ \\
    \midrule
    \texttt{Total} & $288,919$ & $72,661$ \\
    \bottomrule
  \end{tabular}
\end{table}

\section{System overview}

We explored $3$ different ways of laveraging LLM embeddings for either binary or multi-class classification tasks. For subtask $1$ we conducted experiments with systems described in sections \ref{subsec:llm_with_classical_ml} and \ref{subsec:llm_classification_head}. For the second subtask we did our experiments with systems from sections \ref{subsec:llm_classification_head} and \ref{subsec:llm_embeddings}.

\subsection{Frozen LLM embeddings with classical machine learning models}\label{subsec:llm_with_classical_ml}

This approach is using the last layer embeddings extracted from Qwen3-0.6B \cite{yang2025qwen3technicalreport} to train different classic machine learning classifiers:

\begin{itemize}
    \item Light GBM \cite{NIPS2017_6449f44a}
    \item XGBoost \cite{Chen_2016}
    \item Logistic Regression \cite{Berkson01091944}
    \item SVM \cite{708428}
\end{itemize}

In the final approach we used an ensemble of the above models combining predictions in a majority voting manner. So the final prediction is the class voted by most of the models in the ensemble. The embeddings model was kept frozen and used only to extract embeddings from the last layer.

\subsection{LLM fine-tuning with classification head for downstream task}\label{subsec:llm_classification_head}

The next approach also laverages the rich representational capacity of large language models, which can capture both complex linguistic patterns and semantic relationships through their high-dimensional vector space representations, and used as input for the final classification head. Experiments were conducted with embeddings extracted using differnet LLM backbones: Qwen2.5-0.5B \footnote{https://huggingface.co/Qwen/Qwen2.5-0.5B}, Qwen3-0.6B \footnote{https://huggingface.co/Qwen/Qwen3-0.6B}, Mistral7B-v0.1 \footnote{https://huggingface.co/mistralai/Mistral-7B-v0.1}, Llama-3.1-8B \footnote{https://huggingface.co/meta-llama/Llama-3.1-8B} and fine-tuning them with a classification head on top of the last token.

\subsection{Combine transformer-based embeddings at different layers with layer-wise projection and attentive pooling}\label{subsec:llm_embeddings}

We experimented this approach only for subtask 2. It uses embeddings from transformer-based models, combines them using frame-wise projection, pooling strategies and a lightweight classification head.

\subsubsection{Backbone and Feature Extractor}

The model uses a pretrained transformer as a backbone to extract contextualized token embeddings. This part of the model is always kept frozen during the training process.

\subsubsection{Layer Aggregation Block}

The information across multiple hidden layers of the transformer are aggregated using a learnable, softmax-normalized weighted scheme. The weighted sum of hidden states is normalized via layer normalization (without element-wise affine tranformation), producing a single sequence of contextual embeddings for each input.

\subsubsection{Frame-wise Processing Block}

The aggregated embeddings are further processed by a configurable frame-wise module, which can be either a simple linear projection or a small neural network. This step reduces the dimensionality and introduces additional non-linearity, preparing the features for temporal pooling.

\subsubsection{Time Pooling Strategies}

To summarize the sequence of embeddings into a fixed-size representation, the model supports several time pooling strategies:
\begin{itemize}
    \item \textbf{Statistical Pooling (SP):} Concatenates the mean and standard deviation across the sequence.
    \item \textbf{Attentive Statistical Pooling (ASP):} Computes attention-weighted statistics using a learnable attention mechanism.
\end{itemize}

\section{Experimental setup}

For both subtasks, we kept 20\% of the training set for validation and hyperparameter tuning (taking into consedireation the label distribution as well), and the dev split was used to test the final models. For the dataset splits and implementation for classical machine learning models (\ref{subsec:llm_with_classical_ml}) we used scikit-learn \cite{pedregosa2018scikitlearnmachinelearningpython} \footnote{https://scikit-learn.org/stable/}, xgboost \footnote{https://xgboost.readthedocs.io/en/release\_3.0.0/} and lightgbm \footnote{https://lightgbm.readthedocs.io/en/stable/} whereas for LLM fine-tuning (\ref{subsec:llm_classification_head}) and transformer-based embeddings combined (\ref{subsec:llm_embeddings}) we used PyTorch \cite{paszke2019pytorchimperativestylehighperformance} \footnote{https://pytorch.org}.

In order to find the best parameters for each of the classical models from \ref{subsec:llm_with_classical_ml} we did hyperparameter tuning using optuna \cite{akiba2019optunanextgenerationhyperparameteroptimization} \footnote{https://optuna.org} and the final configuration is reported in \autoref{tab:all-hyperparams}.

For the second approach \ref{subsec:llm_classification_head} we loaded the backbone LLMs using QLoRA \cite{dettmers2023qloraefficientfinetuningquantized} (full parameters in \autoref{tab:qlora-params}) in order to be able to fine-tune the entire model (including the LLM backbone) for the downstream task \footnote{All experiments were conducted using 1 Nvidia RTX 4090 GPU}. All hyperparameters used for training in \autoref{tab:llm-finetune-params}.

The backbone transformer model used for the last approach \ref{subsec:llm_embeddings} is RoBERTa \cite{liu2019robertarobustlyoptimizedbert} \footnote{https://huggingface.co/FacebookAI/roberta-base}. The final model we used a hidden dimension of $128$ for the frame-wise processing block and an attentive statistical pooling with $4$ heads and $256$ hidden dimension with ReLU activation. All other training parameters are available in \autoref{tab:roberta-finetune-params}.

The evaluation platform used for this challenge was TIRA \cite{froebe:2023}.

\section{Results}

\subsection{Subtask 1: Voight-Kampff AI Detection Sensitivity}

The metrics used to evaluate the system are the following:

\begin{itemize}
    \item \textbf{ROC-AUC:} The area under the ROC (Receiver Operating Characteristic) curve.
    \item \textbf{Brier:} The complement of the Brier score (mean squared loss).
    \item \textbf{C@1:} A modified accuracy score that assigns non-answers (score = 0.5) the average accuracy of the remaining cases.
    \item \textbf{F1:} The harmonic mean of precision and recall.
    \item \textbf{F0.5u:} A modified F0.5 measure (precision-weighted F measure) that treats non-answers (score = 0.5) as false negatives.
    \item \textbf{Mean:} The arithmetic mean of all the metrics above.
\end{itemize}

The overall results of the first approach \ref{subsec:llm_with_classical_ml} are available in \autoref{table:task1-classicalml} and those using the second approach \ref{subsec:llm_classification_head} are available in \autoref{table:task1-llm-classification-head}.

\begin{figure}[htbp]
    \centering
    \includegraphics[width=1\linewidth]{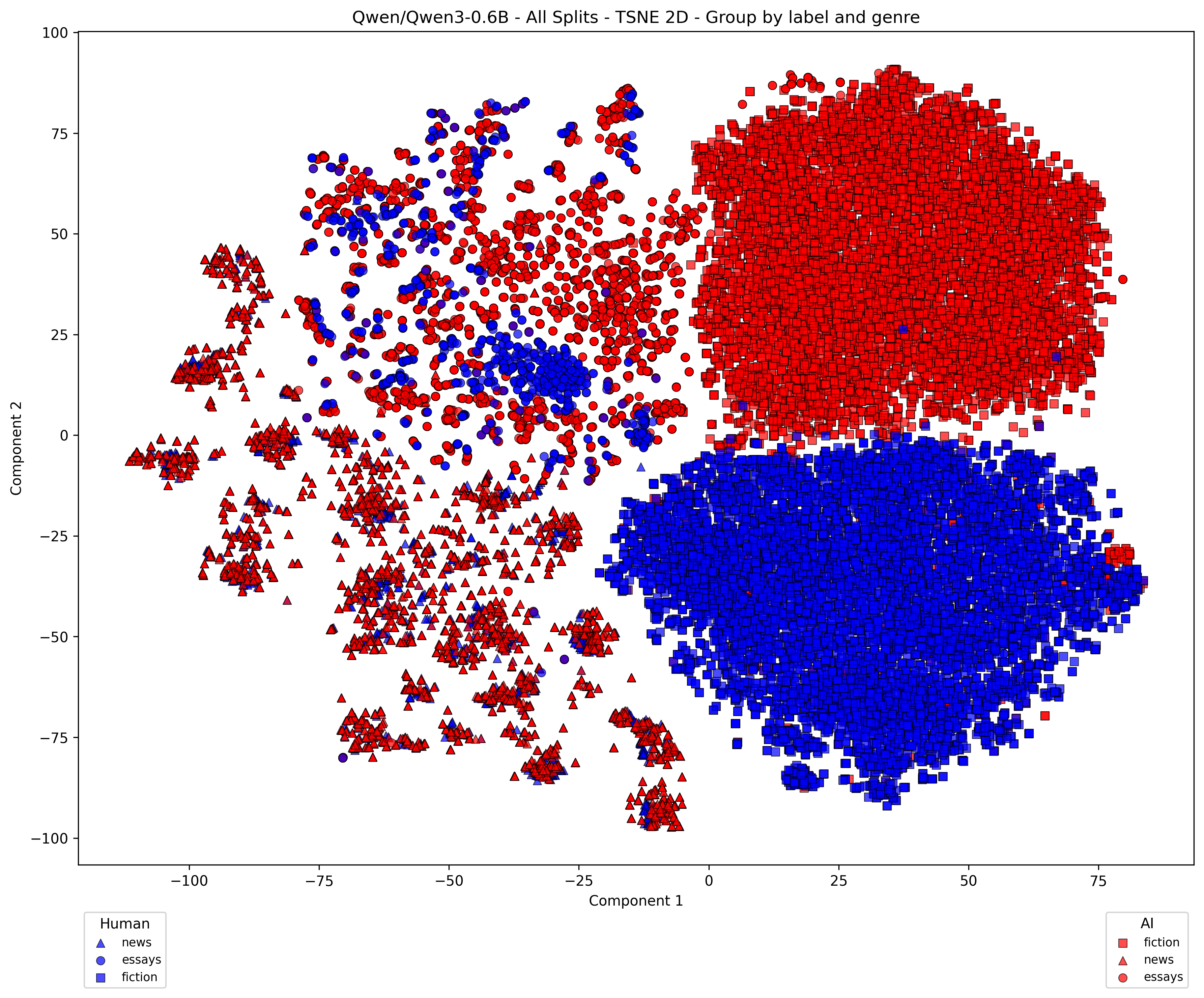}
    \caption{Qwen3-0.6B embeddings visualized in 2D using T-SNE (tarin + dev)}
    \label{fig:task1-embeddings-visualization}
\end{figure}

\begin{figure}[htbp]
    \centering
    \includegraphics[width=1\linewidth]{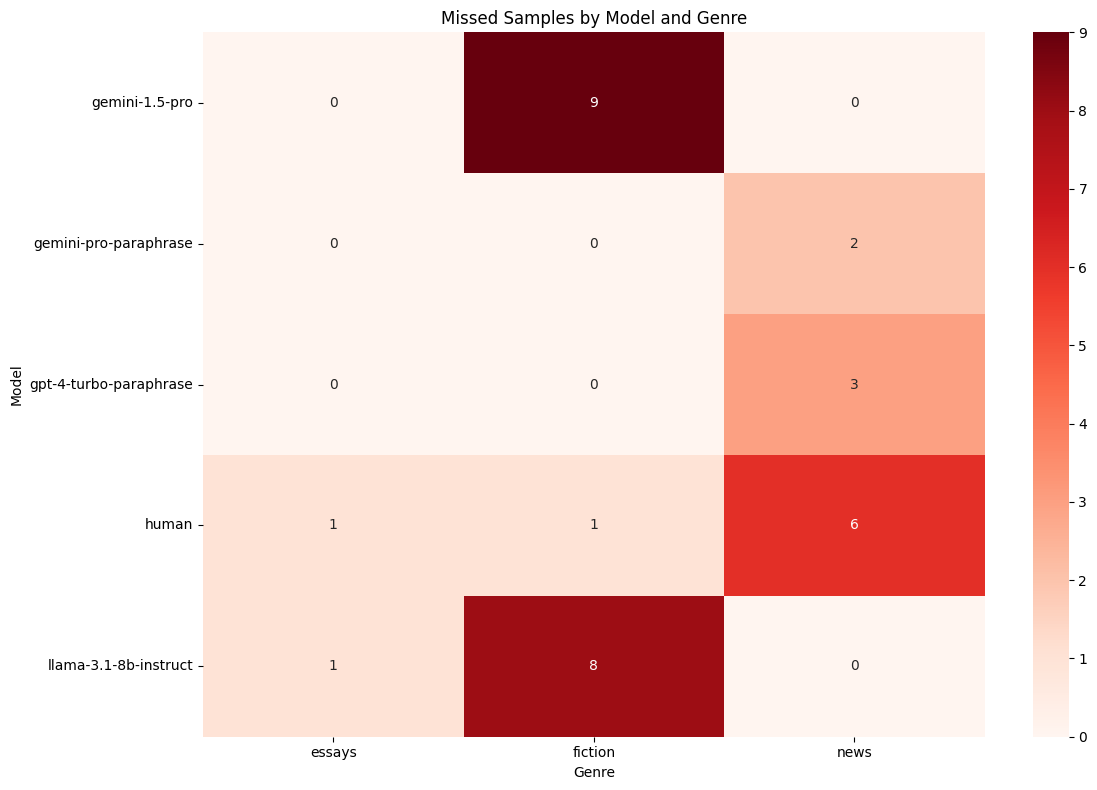}
    \caption{Missed samples on the dev set by the ensemble model for subtask 1.}
    \label{fig:task1-heatmap}
\end{figure}

Looking at \autoref{fig:task1-heatmap} it seems like the biggest problem for the model is that is classifies fiction texts generated with gemini-1.5-pro and llama-3.1-8b-instruct as human-written fiction texts. In the embeddings visualization plot \autoref{fig:task1-embeddings-visualization} we can see that there are some red squares (ai-written fiction) that are really close to the blue squares (human-written fiction).

\begin{table}[ht]
\centering

\caption{Reported metrics for the classical ML models on the dev set.}
\label{table:task1-classicalml}

\begin{tabular}{lcccccc}
\hline
Model & ROC-AUC & Brier & C@1 & F1 & F0.5u & Mean \\
\hline
XGBoost & 0.999 & 0.991 & 0.989 & 0.991 & 0.992 & 0.992 \\
Light GBM & 0.999 & 0.992 & 0.989 & 0.991 & 0.992 & 0.993 \\
Logistic Regression & 0.999 & 0.994 & 0.994 & 0.995 & 0.996 & 0.996 \\
SVM & 0.999 & 0.994 & 0.992 & 0.994 & 0.996 & 0.995 \\
Ensemble & 0.999 & 0.994 & 0.991 & 0.993 & 0.995 & 0.995 \\
\hline
\end{tabular}
\end{table}

\begin{table}[ht]
\centering

\caption{Reported metrics for the fine-tuning LLM approach on the dev set.}
\label{table:task1-llm-classification-head}

\begin{tabular}{lcccc}
\hline
Model & F1 & Accuracy & Precision & Recall \\
\hline
Qwen2.5-0.5B & 0.978 & 0.972 & 0.980 & 0.977 \\
Qwen3-0.6B   & 0.993 & 0.991 & 0.993 & 0.993 \\
Llama-3.1-8B & 0.997 & 0.996 & 0.999 & 0.996 \\
Mistral-7B-v0.1 & 0.995 & 0.993 & 0.997 & 0.992 \\
\hline
\end{tabular}

\end{table}

\subsection{Subtask 2: Human-AI Collaborative Text Classification}

The metrics used to evaluate the system are the following:

\begin{itemize}
    \item \textbf{Recall (Macro):} Macro recall is the average recall computed independently for each class, measuring the proportion of true positives identified out of all actual instances for each class, and then averaging across all classes.
    \item \textbf{F1 (Macro):} Macro F1 is the average of the F1 scores calculated for each class, where the F1 score is the harmonic mean of precision and recall, providing a balanced measure of a model’s accuracy across all classes.
    \item \textbf{Accuracy:} Accuracy is the proportion of all predictions that are correct, calculated as the number of correct predictions divided by the total number of predictions.
\end{itemize}

The overall results of the third approach \ref{subsec:llm_embeddings} are available in \autoref{table:task2-llm-embeddings}.

\begin{table}[ht]
\centering

\caption{Reported metrics for the combined transformer-based embeddings using layer-wise projection and attentive pooling on dev and test sets.
*Based on the error analysis \autoref{fig:task2-heatmap-source-true_label} we reported the metrics on dev set removing the samples from m4gt-bench which proved to be the harderst to predict.}
\label{table:task2-llm-embeddings}

\begin{tabular}{l|ccc|r}
\hline
Model & Recall (Macro) & F1 (Macro) & Accuracy & Split\\
\hline
RoBERTa-base & 0.750 & 0.592 & 0.541 & Dev \\
RoBERTa-base & 0.443 & 0.427 & 0.514 & Test \\
\hline
RoBERTa-base & 0.890 & 0.842 & 0.917 & Dev (no samples from m4gt-bench)* \\
\hline
\end{tabular}
\end{table}

\begin{figure}[htbp]
    \centering
    \includegraphics[width=1\linewidth]{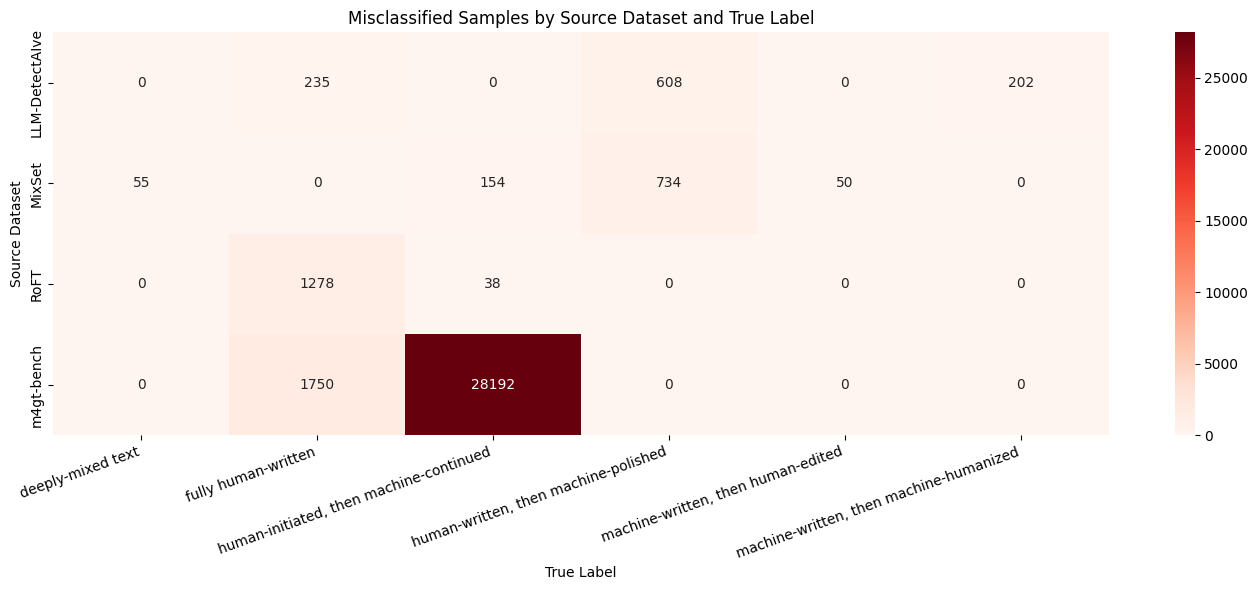}
    \caption{Heatmap of misclassified samples for each source dataset on dev set subtask 2.}
    \label{fig:task2-heatmap-source-true_label}
\end{figure}

\begin{figure}[htbp]
    \centering
    \includegraphics[width=1\linewidth]{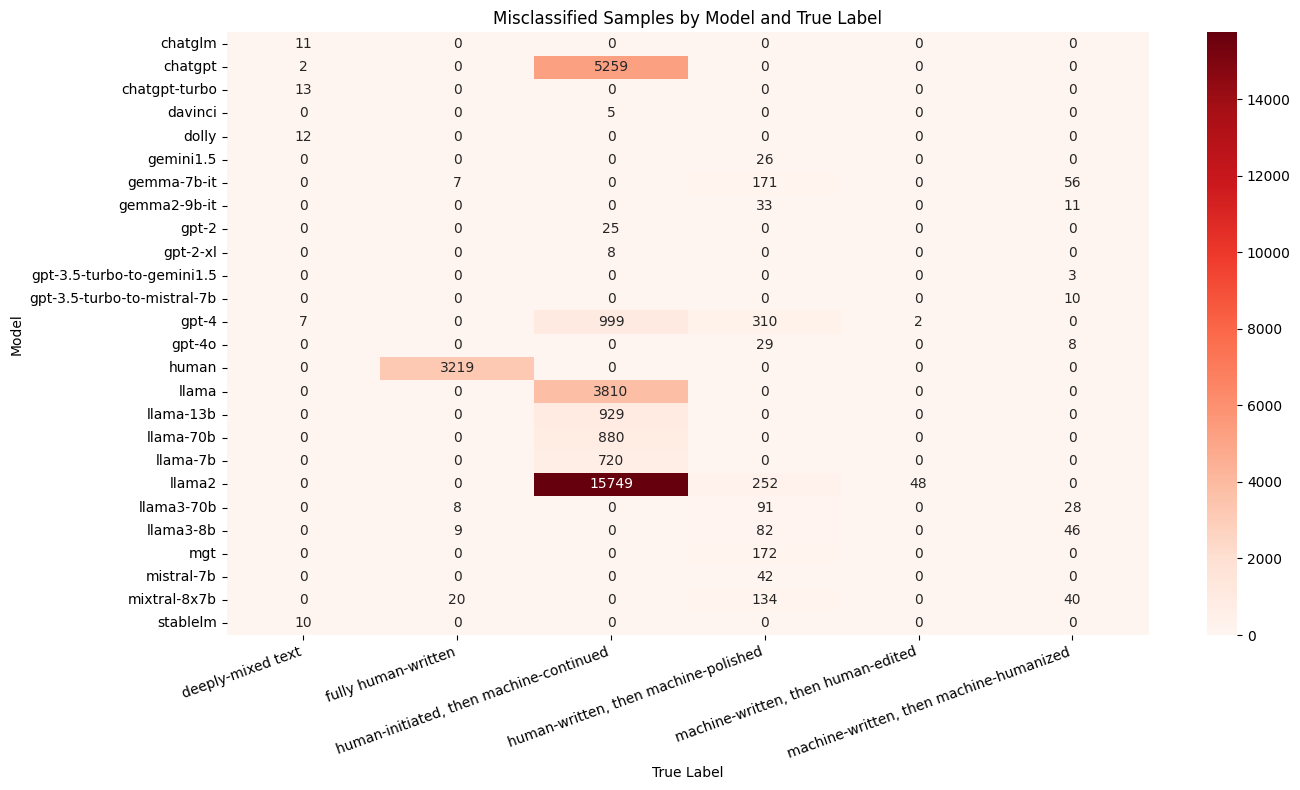}
    \caption{Heatmap of misclassified samples for each model on dev set subtask 2.}
    \label{fig:task2-heatmap-model-true_label}
\end{figure}

\begin{figure}[htbp]
    \centering
    \includegraphics[width=1\linewidth]{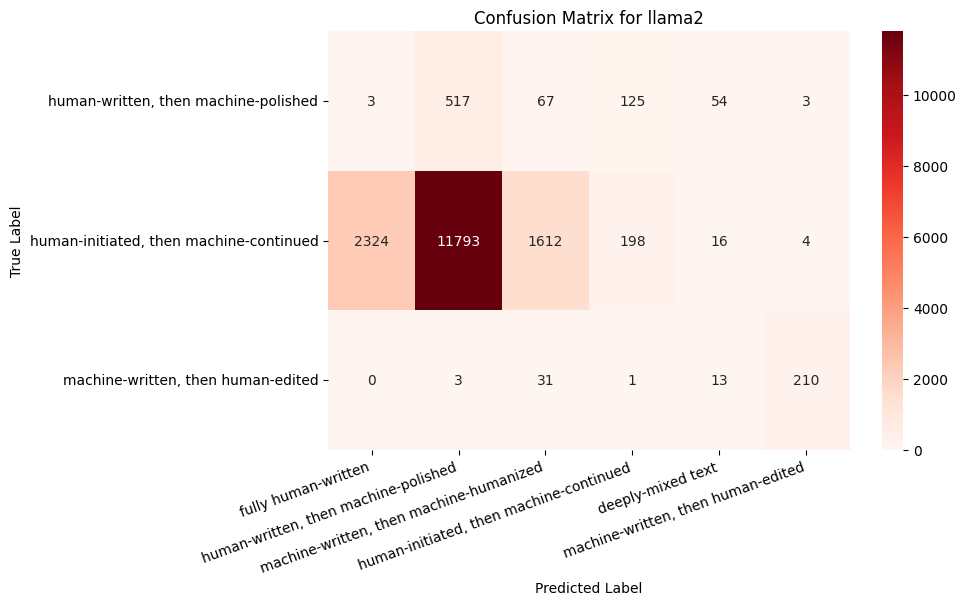}
    \caption{Confusion matrix of samples generated using llama2 model on dev set subtask 2.}
    \label{fig:task2-confusion-matrix-llama}
\end{figure}

Looking at \autoref{fig:task2-heatmap-source-true_label} we can see that the hardest dataset is m4gt-bench (which was not present in the training dataset at all) with $28,192$ samples that were not classified as human-initiated, then machine-continued ($97\%$ of the samples from m4gt-bench with this label) and $1,750$ were not classifier as fully human-written ($55\%$ of the samples from m4gt-bench with this label). Furthermore looking at \autoref{fig:task2-heatmap-model-true_label} and \autoref{fig:task2-confusion-matrix-llama} we see that llama2 is the hardest model to be detected (also not present in the training split at all) with $11,793$ samples that were human-written, then machine-continued misclassified as human-written, then machine-polished. We have also reported the metrics we got on the dev set when removing the samples from m4gt-bench \autoref{table:task2-llm-embeddings} to prove that the architecture we used for this type of classification fails to generalize well on unseen data, but manages to learn good enough representations from the training set.

\section{Conclusions and Future Work}

The experimental results demonstrate that both classical machine learning models and fine-tuned large language models (LLMs) achieve exceptionally high performance on the subtask 1 of distinguishing between human-written and ai-written texts, with F1, ROC-AUC, and related metrics consistently above 0.99 for most models. However, detailed error analysis reveals that the main challenge lies in distinguishing between human-written and AI-generated fiction, especially for texts produced by advanced models such as gemini-1.5-pro and llama-3.1-8b-instruct, which were also fewer in the train set, where the model tends to misclassify AI-generated fiction as human-written. This is further supported by the embeddings visualization, which shows significant overlap between these classes.

For the human-AI collaborative text classification subtask, the results are more modest, with macro F1 and recall scores around 0.43 and accuracy at 0.51, suggesting that this remains a challenging problem and that current models struggle to generalize in this setting, the difference between human-written then machine-continued and human-written then machine-polished being very subtle and hard to be detected, especially when the generative model has never been seen in the training dataset.

Future work will focus on addressing these limitations by exploring more sophisticated representation learning techniques, incorporating additional context or metadata, and experimenting with advanced ensemble methods. Furthermore, targeted data augmentation and adversarial training could help the models better distinguish subtle differences between human and machine-generated texts, particularly in challenging genres such as fiction. Finally, interpretability analyses and error-driven model refinement will be essential to further improve robustness and reliability in real-world applications.

\section*{Acknowledgements}

This research is supported by the project “Romanian Hub for Artificial Intelligence - HRIA”, Smart Growth, Digitization and Financial Instruments Program, 2021-2027, MySMIS no. 334906.



\bibliography{participant-lit}

\appendix

\section*{Declaration on Generative AI}
 During the preparation of this work, the authors used ChatGPT in order to: Generate literature review, Grammar and spelling check. After using these tools, the authors reviewed and edited the content as needed and take full responsibility for the publication’s content.

\begin{table}[ht]
\centering
\caption{Best hyperparameters for classical models used in the final ensemble model obtained using optuna.}
\label{tab:all-hyperparams}
\begin{tabular}{lcccc}
\hline
Hyperparameter & XGBoost & LightGBM & Logistic Regression & SVM \\
\hline
n\_estimators      & 260    & 218    & --     & --     \\
learning\_rate     & 0.1114 & 0.2004 & --     & --     \\
max\_depth         & 8      & 9      & --     & --     \\
num\_leaves        & --     & 37     & --     & --     \\
subsample          & 0.8541 & 0.9007 & --     & --     \\
colsample\_bytree  & 0.8348 & 0.9641 & --     & --     \\
min\_child\_weight & 4      & --     & --     & --     \\
min\_child\_samples& --     & 69     & --     & --     \\
reg\_alpha         & --     & 0.4093 & --     & --     \\
reg\_lambda        & --     & 0.6594 & --     & --     \\
gamma              & 0.0913 & --     & --     & --     \\
C                  & --     & --     & 0.1156 & 2.5501 \\
penalty            & --     & --     & l1     & l1     \\
solver             & --     & --     & saga   & --     \\
max\_iter          & --     & --     & 524    & 1229   \\
random\_state      & --     & --     & 42     & --     \\
loss               & --     & --     & --     & squared\_hinge \\
tol                & --     & --     & --     & 0.00015 \\
\hline
\end{tabular}
\end{table}


\begin{table}[ht]
\centering
\caption{QLoRA-related hyperparameters used for all fine-tuned LLMs.}
\label{tab:qlora-params}
\begin{tabular}{lc}
\hline
Parameter & Value \\
\hline
load\_in\_4bit      & True \\
load\_in\_8bit      & False \\
use\_double\_quant  & True \\
quant\_type         & nf4 \\
compute\_dtype      & bfloat16 \\
lora\_r             & 8 \\
lora\_alpha         & 32 \\
lora\_dropout       & 0.05 \\
lora\_bias          & none \\
lora\_target\_modules & \texttt{\{q\_proj, k\_proj, v\_proj, o\_proj\}} \\
\hline
\end{tabular}
\end{table}


\begin{table}[ht]
\centering
\caption{Fine-tuning hyperparameters used for all LLMs.}
\label{tab:llm-finetune-params}
\begin{tabular}{lc}
\hline
Parameter & Value \\
\hline
batch\_size                & 16 \\
num\_epochs                & 0.5 \\
learning\_rate             & 2e-5 \\
lr\_scheduler\_type        & constant \\
max\_grad\_norm            & 0.3 \\
max\_length                & 1024 \\
weight\_decay              & 0.01 \\
warmup\_ratio              & 0.03 \\
optim                      & paged\_adamw\_32bit \\
gradient\_accumulation\_steps & 4 \\
eval\_strategy             & steps \\
save\_strategy             & steps \\
seed                       & 42 \\
\hline
\end{tabular}
\end{table}


\begin{table}[ht]
\centering
\caption{Training hyperparameters for RoBERTa-base fine-tuning.}
\label{tab:roberta-finetune-params}
\begin{tabular}{lc}
\hline
Parameter & Value \\
\hline
model\_name                & FacebookAI/roberta-base \\
batch\_size                & 4 \\
num\_epochs                & 5 \\
learning\_rate             & 0.0003 \\
max\_length                & 512 \\
gradient\_accumulation\_steps & 4 \\
seed                       & 42 \\
\hline
\end{tabular}
\end{table}



\end{document}